\documentclass[sigconf]{acmart}

\usepackage{booktabs} 

\setcopyright{none}

\acmDOI{}

\acmISBN{}

\acmConference[]{}{}{} 
\acmYear{}
\acmPrice{}
\usepackage{graphicx}
\usepackage{amsmath,amssymb}
\usepackage[titletoc,title]{appendix}

\usepackage{etoolbox}
\newtoggle{reveal_id}
\toggletrue{reveal_id}

\newcommand{\ma}[1]{\textbf{#1}}
\newcommand{\ve}[1]{\textbf{#1}}
\newcommand{\se}[1]{\mathsf{#1}}

\usepackage[noend]{algorithmic}
\usepackage{algorithm}

\algsetup{indent=0.5cm}

\usepackage{color}

\begin{document}


\title{Getting Deep Recommenders Fit: Bloom Embeddings for Sparse Binary Input/Output Networks}




\iftoggle{reveal_id}{
\author{Joan Serr{\`a} and Alexandros Karatzoglou}
\affiliation{%
  \institution{Telef{\'o}nica Research}
  \streetaddress{Pl.~Ernest Lluch i Mart{\'i}n, 5}
  \city{Barcelona} 
  \state{Spain} 
  \postcode{08019}
}
\email{firstname.lastname@telefonica.com}
}



\begin{abstract}
Recommendation algorithms that incorporate techniques from deep learning are becoming increasingly popular. Due to the structure of the data coming from recommendation domains (i.e.,~one-hot-encoded vectors of item preferences), these algorithms tend to have large input and output dimensionalities that dominate their overall size. This makes them difficult to train, due to the limited memory of graphical processing units, and difficult to deploy on mobile devices with limited hardware. To address these difficulties, we propose Bloom embeddings, a compression technique that can be applied to the input and output of neural network models dealing with sparse high-dimensional binary-coded instances. Bloom embeddings are computationally efficient, and do not seriously compromise the accuracy of the model up to 1/5 compression ratios. In some cases, they even improve over the original accuracy, with relative increases up to 12\%. We evaluate Bloom embeddings on 7~data sets and compare it against 4~alternative methods, obtaining favorable results. We also discuss a number of further advantages of Bloom embeddings, such as `on-the-fly' constant-time operation, zero or marginal space requirements, training time speedups, or the fact that they do not require any change to the core model architecture or training configuration.
\end{abstract}

%
%





\maketitle


\section{Introduction}
\label{sec:intro}

The size of neural network models that deal with sparse inputs and outputs is often dominated by the dimensionality of such inputs and outputs. Deep networks used for recommender systems and collaborative filtering are a paradigmatic case, as they have to deal with high-dimensional sparse vectors, typically in the order from tens of thousands to hundreds of millions, both at the input and the output of the network (e.g.,~\cite{Wu2016,Hidasi16ICLR,Cheng:2016,Strub:2016}). This results in large models that present a number of difficulties, both at training and prediction stages. Apart from training and prediction times, an obvious bottleneck of such models is space: their size (and even performance) is hampered by the physical memory of graphical processing units (GPUs), and they are difficult to deploy on mobile devices with limited hardware (cf.~\cite{Han16ICLR}).

One option to reduce the size of sparse inputs and outputs is to embed them into a lower-dimensional space. Embedding sparse high-dimensional inputs is commonplace (e.g.,~\cite{Bengio00NIPS,Turian10ACL,Mikolov13ARXIV}). However, embedding sparse high-dimensional outputs, or even inputs and outputs at the same time, is much less common (cf.~\cite{Weston02NIPS,Bengio10NIPS,Akata15PAMI}). Importantly, typical embeddings still require the storage and processing of large matrices with the same dimensionality as the input/output (like the original neural network model would do). Thus, the gains in terms of space are limited. As mentioned, the size of such models is dominated by the input/output dimensionality, with input and output layers representing around 99.9\% of the total amount of models' parameters. An example can be found in the deep recommender of Hidasi et~al.~\cite{Hidasi16ICLR}, which uses a gated recurrent unit to perform session-based recommendations with input/output layers of dimensionality 330,000 and internal layers of dimensionality 100.

In general, an ideal embedding procedure for sparse high-\linebreak dimensional inputs/outputs should produce compact embeddings, of much lower dimensionality than the original input/output. In addition, it should consume little space, both in terms of storage and memory space. Smaller sizes imply less parameters, thus training the model on embedded vectors would also be faster than with the original instances. The embedding of the output should also lead to a formulation for which the appropriate loss should be clear. Embeddings should not compromise the accuracy of the model nor the required number of training epochs to obtain that accuracy. In addition, no changes to the original core architecture of the model should be required to achieve good performance (obviously, input/output dimensions must change). The embedding should also be fast; if not to be done directly `on-the-fly', at least fast enough so that speed improvements made during training  are not lost in the embedding operation. Last, but not least, output embeddings should be easily reversible, so that the output of the model could be mapped to the original items at prediction time.

In this work, we propose an unsupervised embedding technique that fulfills all the previous requirements. It can be applied to both input and output layers of neural network models that deal with binary (one-hot encoded) inputs and/or outputs. In doing so, it produces lower-dimensionality binary embeddings that can be easily mapped to the original instances. Provided that the embedding dimension is not too low, the accuracy is not compromised. Furthermore, in some cases, we show that training with embedded vectors can even increase prediction accuracy (we find that is the case for three of the considered recommendation tasks). The embedding requires no changes to the core network structure nor to the model configuration, and works with a softmax output, the most common output activation for binary-coded instances. The embedding moreover preserves the ranking order of items which is crucial in recommender systems.  As it is unsupervised, the embedding does not require any preliminary training. Moreover, it is a constant-time operation that can be either performed on-the-fly, requiring no disk or memory space, or can be cached in memory, occupying orders of magnitude less space than a typical embedding matrix. Lower dimensionality of input/output vectors result in faster training, and the mapping from the embedded space to the original one does not add an overwhelming amount of time to the prediction stage. The proposed embedding is based on the idea of Bloom filters~\cite{Bloom70CACM}, and therefore it inherits part of the theory developed around that idea~\cite{Blustein02TR,Dillinger04FMCAD,Mitzenmacher05BOOK,Bonomi06ESA}.


\section{Related work}
\label{sec:relatedwork}

A common approach to embed high-dimensional inputs is the hashing trick~\cite{Langford07TECHREP,Shi09JMLR,Weinberger09ICML}. However, the hashing trick approach does not deal with outputs, as it offers no explicit way to map back from the (dense) embedding space to the original space. A more elementary version of the hashing trick~\cite{Ganchev08MLP} can be used at the outputs by considering it as a special case of the Bloom-based methodology proposed here. A framework providing both encoding and decoding strategies is the error-correcting output codes (ECOC) framework~\cite{Dietterich95JAIR}. Originally designed for single-class outputs, it can be also applied to class sets~\cite{Armano12IIR}. Another example of a framework offering recovery capabilities is kernel dependency estimation~\cite{Weston02NIPS}. The compressed sensing approach of Hsu et~al.~\cite{Hsu09NIPS} builds on top of ECOC to reduce multi-label regression to binary regression problems. Similarly, Ciss{\'e} et~al.~\cite{Cisse13NIPS} use Bloom filters to reduce multi-label classification to binary classification problems and improve the robustness of individual binary classifiers' errors. 

Data-dependent embeddings that require some form of learning also exist. A typical approach is to rely on variants of latent semantic analysis or singular value decomposition (SVD), exploiting similarities or correlations that may be present in the data. Again, the issue of mapping from the embedding space to the original space is left unresolved. Nonetheless, recently, Chollet~\cite{Chollet16ARXIV} has successfully applied a K-nearest neighbors (KNN) algorithm to perform such a mapping and to derive a ranking of the items in the original space. An SVD decomposition of the pairwise mutual information matrix (PMI) is used to perform the embedding, and cosine similarity is used as loss function and to retrieve neighbors. Using the KNN trick offers the possibility to exploit different types of factorization of similarity-based matrices. Canonical correlation analysis is an example that considers both inputs and outputs at the same time~\cite{Hotelling36BM}. Other examples considering output embeddings are nuclear norm regularized learning~\cite{Amit07ICML}, label embedding trees~\cite{Bengio10NIPS}, or the WSABIE algorithm~\cite{Weston10ML}.
In the presence of side information, like text descriptions, item or class taxonomies, or manually-collected data, a range of approaches are applicable. Akata et~al.~\cite{Akata15PAMI} provide a comprehensive list. In our study, we assume no side information is available and focus on input/output-based embeddings.

From a more general perspective, reducing the space of (or compressing) neural network models is an active research topic, driven by the need to deploy such models in systems with limited hardware resources. A common approach is to reduce the size of already trained models by some quantization and/or pruning of the connections in dense layers~\cite{Courbariaux15NIPS,Han16ICLR,Kim16ICLR}. A less frequent approach is to reduce the model size before training~\cite{Chen15ICML}. These methods typically do not focus on input layers and, to the best of our knowledge, none of them deals with high-dimensional outputs.
It is also worth noting that a number of techniques have been proposed to efficiently deal with high-dimensional outputs, specially in the natural language processing domain. The hierarchical softmax approach~\cite{Morin05AISTATS} or the more recent adaptive softmax~\cite{Grave16ARXIV} are two examples of those. Yet, as mentioned, the focus of these works is on speed, not on space. The work of Vincent et~al.~\cite{Vincent15NIPS} focuses on both aspects of very large sparse outputs but, to the best of our knowledge, cannot be directly applied to common softmax outputs.


\section{Bloom embeddings}
\label{sec:bloom}

\subsection{Bloom filters}
\label{sec:bloom_bf}

Bloom filters~\cite{Bloom70CACM} are a compact probabilistic data structure that is used to represent sets of items, and to efficiently check whether an item is a member of a set~\cite{Mitzenmacher05BOOK}. Since the instances we deal with represent sets of one-hot encoded items, Bloom filters are an interesting option to embed those in a compact space with good recovery (or checking) guarantees. 

In essence, Bloom filters project every item of a set to $k$ different positions of a binary array $\ve{u}$ of size $m$. Projections are done using a set of $k$ independent hash functions $\se{H}=\left\{H_i\right\}_{i=1}^{k}$, each of which with a range from 1 to $m$, ideally distributing the projected items uniformly at random~\cite{Mitzenmacher05BOOK}. Proper independent hash functions can be derived using enhanced double hashing or triple hashing~\cite{Dillinger04FMCAD}. The number of hash functions $k$ is usually a constant, $k\ll m$, proportional to the expected number of items to be projected.  

To check if an item is in $\ve{u}$, one feeds it to the $k$ hash functions $H$ to get $k$ array positions. If any of the bits at these positions is 0, then the item is definitely not in the set. Thus, item checks return no false negatives, meaning that the structure gives an answer with 100\% recall~\cite{Mitzenmacher05BOOK}. However, if all $k$ bits at the projected positions are 1, then either the item is in the set, or the bits have by chance been set to 1 during the insertion of other set items. This implies that false positives are possible, due to collisions between projections of different items~\cite{Blustein02TR}. The values of $m$ and $k$ can be adjusted to control the probability of such collisions. However, in practice, $m$ is usually constrained by space requirements, and $k\leq 10$ is employed, independently of the number of items to be projected, and giving less than 1\% false positive probability~\cite{Bonomi06ESA}.


\subsection{Embedding and recovery}
\label{sec:bloom_be}

In the following, we describe the use of Bloom filter techniques in embedding binary high-dimensional instances, and the recovery or mapping to such instances from these embeddings. We denote our approach as Bloom embedding (BE). The idea we pursue is to embed both inputs and outputs and to perform training in the embedding space. For that, only a probability-based output activation is required, together with a loss function that is appropriate for such activations. 

Let $\ve{x}$ be an input or output instance with dimensionality $d$, such that $\ve{x}=\left[x_1,\dots x_d\right]$, $x_i\in\{0,1\}$. Instances $\ve{x}$ are assumed to be sparse, that is, $\sum_{i=1}^{d} x_i\ll d$. Because of that, we can more conveniently (and compactly) represent $\ve{x}$ as set $\se{p}=\left\{ p_i\right\}_{i=1}^{c}$, $p_i\in\mathbb{N}_{\leq d}$, where $c$ is the number of non-zero elements and $p_i$ is the position of such elements in $\ve{x}$. For every set $\se{p}$, we generate an embedded instance $\ve{u}$ of dimensionality $m<d$, such that $\ve{u}=\left[u_1,\dots u_m\right]$, $u_i\in\{0,1\}$. To do so, we first set all $m$ components of $\ve{u}$ to 0. Then, iteratively, for every position $p_i$, $i=1,\dots c$, and every projection $H_j$, $j=1,\dots k$, we assign
\begin{equation}
u_{H_j(p_i)} = 1 .
\label{eq:assign}
\end{equation}

Notice that, since $H_j$ has a range between 1 and $m$, $k\ge 1$, and $m<d$, a number of original positions $p_i$ may map to the same index in $\ve{u}$. Bloom filters mitigate this by properly choosing $k$ independent hash functions $H$ (Sec.~\ref{sec:bloom_bf}). Notice furthermore that the process has no space requirements, as $H$ is computed on-the-fly. Finally, notice that the embedding of a set $\se{p}$ is constant time: the process is $O(ck)$, with $c$ bounded by the maximum number of active items in $\ve{x}$, $c\ll d$, and $k$ being a constant that is set beforehand, $k\ll m<d$. In practice, this constant time is dominated by the time spent on $H$ to generate a hash. If we want to be faster than that, and at the same time ensure an optimal (uniform) distribution of the outputs of $H$, we can pre-compute a hash matrix storing the projections or hash indices for all the  potential items in $\se{p}$. We can do it by generating vectors $\ve{h}=\left[h_1,\dots h_k\right]$ for each $p_i$, where $h_j$ is a uniformly randomly chosen integer between 1 and $m$ (without replacement). This way, by pre-generating all projections for all $d$ items, we end up with a $d\times k$ matrix $\ma{H}$ of integers between 1 and $m$, which we can easily store in random-access memory (RAM), not in the GPU memory. 

We now explain how to recover a probability-based ranking of the $d$ items at the output of the model. Assuming a softmax activation is used, we have a probability vector $\hat{\ve{v}}=\left[\hat{v}_1,\dots \hat{v}_m\right]$ that, at training time, is compared to the binary embedding $\ve{v}$ of the ground truth $\ve{y}$, with its active positions $\se{q}$ (analogously to $\ve{x}$ and $\se{p}$). 
To unravel the embedding $\hat{\ve{v}}$ and map to the $d$ original items of $\ve{y}$, we can understand $\hat{\ve{v}}$ as a $k$-way factorization of every item $\hat{y}_i$ of our prediction $\hat{\ve{y}}$. Then, following the idea of Bloom filters, if $y_i$ maps to $v_i$ and $\hat{v}_i=0$, we can confirm that that item is definitely not in the output of the model (Sec.~\ref{sec:bloom_bf}). Otherwise, if $\hat{v}_i$ is relatively large, we want the likelihood of that item to reflect that.
Specifically, given an active position $q_i$ from $\se{q}$ representing $\ve{y}$, we can compute the likelihood
\begin{equation}
L(q_i) = \prod_{j=1}^{k} \hat{v}_{H_j(q_i)} ,
\label{eq:like1}
\end{equation}
and assign outputs $\hat{y}_i=L(q_i)$. Alternatively, if a more numerically-stable output is desired, we can compute the negative log-likelihood
\begin{equation}
L(q_i) = -\sum_{j=1}^{k} \log\left(\hat{v}_{H_j(q_i)}\right) .
\label{eq:like2}
\end{equation}
Both operations, when iterated for $i=1,\dots d$, define a ranking over the items in $\ve{y}$ and from there, if needed, a probability distribution can be recovered by re-normalization. We here do not perform such re-normalization, as all the problems we consider can be mapped to a typical ranking recommendation setting.

\subsection{Suitability}
\label{sec:bloom_suitable}

Note that BE, by construction, already offers a number of the aforementioned desired qualities for sparse binary high-dimensional embeddings (Sec.~\ref{sec:intro}). Specifically, BE is designed for both inputs and outputs, offering a rank-based mapping between the original instances and the embedded vectors. BE yields a more compact representation of the original instance and requires no disk or memory space (at most some marginal RAM space, not GPU memory). In addition, BE can be performed on-the-fly, without training, and in constant time. In the following, we demonstrate the remaining desirable qualities using a comprehensive experimental setup: we show that the accuracy of the model remains stable or even increases given a reasonable embedding dimension, that no changes in the model architecture nor configuration are required, that training times are faster thanks to the reduction of the number of parameters of the model, that evaluation times do not carry much overhead, and that performance is generally better than a number of alternative approaches.


\section{Experimental setup}
\label{sec:setup}

\subsection{General considerations}
\label{sec:setup_consider}

We demonstrate that BE works under several settings and that it can be applied to multiple tasks. In particular, we focus on recommendation and collaborative filtering but also demonstrate the validity of the approach on a natural language processing task. We consider a number of data sets, network architectures, configurations, and evaluation measures. In total, we define 7~different setups, which we describe in Sec.~\ref{sec:setup_data}. We also demonstrate that BE is competitive with respect to the available alternatives. To this end, we consider 4~different state-of-the-art approaches, which we overview in Sec.~\ref{sec:setup_altern}. 

Data sets are formed by inputs with $n$ instances, corresponding to either individual instances (or one-hot encoded user profiles) or to sequences of instances (or profile lists). Outputs, also of $n$ instances, correspond to individual instances or to class labels. Instances have an original dimensionality $d$, corresponding to the cardinality of all possible profile items. Given the nature of the considered problems, instances are very sparse, with all but $c$ items being different from 0, $c\ll d$, typically with $c/d$ in the order of $10^{-5}$ (Table~\ref{tab:data}).

\begin{table}[t]
    \caption{Data set statistics after data cleaning and splitting. From left to right: data set name, number of instances $n$, test split size, instance dimensionality $d$, median number of non-zero components $c$, and median density $c/d$.}
    \label{tab:data}
    \setlength{\tabcolsep}{7pt}
    \begin{tabular}{lrrrcc}
        \toprule
        Data set & \multicolumn{1}{c}{$n$} & \multicolumn{1}{c}{Split} & \multicolumn{1}{c}{$d$}      & $c$     & $c/d$ \\
        \midrule
        ML      & 138,224   & 10,000 & 15,405    & 18            & $1.2\cdot10^{-3}$ \\
        PTB     & 929,589   & 82,430 & 10,001    & ~\,1             & $1.0\cdot10^{-4}$ \\
        CADE    & 40,983    & 13,661 & 193,998   & 17            & $8.8\cdot10^{-5}$ \\
        MSD     & 597,155   & 50,000 & 69,989    & ~\,5             & $7.1\cdot10^{-5}$ \\
        AMZ     & 916,484   & 50,000 & 22,561    & ~\,1             & $4.4\cdot10^{-5}$ \\
        BC      & 25,816    & 2,500 & 54,069    & ~\,2             & $3.7\cdot10^{-5}$ \\
        YC      & 1,865,997 & 50,000 & 35,732    & ~\,1             & $2.8\cdot10^{-5}$ \\
        \bottomrule
    \end{tabular}
\end{table}

For each data set, and based on the literature, we select an appropriate baseline neural network architecture. We experiment with both feed-forward (autoencoder-like) and recurrent networks, carefully selecting their parameters and configuration to match (or even improve) the state-of-the-art results. For the sake of comparison, we also choose appropriate and well-known evaluation measures~\cite{Manning08BOOK}. Depending on the data set (Table~\ref{tab:baselines}), we work with mean average precision (MAP), reciprocal ranks (RR), or \% of accuracy (Acc). 

\begin{table}[t]
    \caption{Experimental setup and baseline scores. From left to right: data set name, network architecture and optimizer, evaluation measure, random score $S_{\text{R}}$, and baseline score $S_0$.}
    \label{tab:baselines}
    \begin{tabular}{lllcc}
        \toprule
        Data set & Architecture + Optimizer & Meas. & $S_{\text{R}}$ & $S_0$ \\
        \midrule
        ML & Feed-forward + Adam & MAP & 0.003 & 0.160 \\
        PTB & LSTM + SGD & RR & 0.001 & 0.342 \\
        CADE & Feed-forward + RMSprop & Acc & 8.5 & 58.0 \\
        MSD & Feed-forward + Adam & MAP & $<$0.001 & 0.066 \\
        AMZ & Feed-forward + Adam & MAP & $<$0.001 & 0.049 \\
        BC & Feed-forward + Adam & MAP & $<$0.001 & 0.010 \\
        YC & GRU + Adagrad & RR & $<$0.001 & 0.368 \\
        \bottomrule
    \end{tabular}
\end{table}

Each combination of data set, network architecture, configuration, and evaluation measure defines a task. For every task, we compute a baseline score $S_0$, corresponding to running the plain neural network model without any embedding. We then report the performance of the $i$-th execution with a particular embedding with respect to the baseline score using $S_i/S_0$. This way, we can compare the performance across different tasks using different evaluation measures, reporting relative improvement/loss with respect to the baseline. Similarly, to compare across different dimensionalities, we report the ratio of embedding dimensionality with respect to the original dimensionality, $m/d$, and to compare across different training and evaluation times, we report time ratios with respect to the baseline, $T_i/T_0$.

\subsection{Tasks}
\label{sec:setup_data}

We now give a brief summary of the 7~considered tasks (Tables~\ref{tab:data} and~\ref{tab:baselines}). All data sets are publicly-available and, for all tasks, we make sure that the network architecture and the experimental setup is sufficient to achieve a state-of-the-art result. We use softmax outputs and categorical cross-entropy losses in all experiments.

\begin{enumerate}

\item Movielens (ML): movie recommendation with the Movielens 20M data set\footnote{\url{http://grouplens.org/datasets/movielens/20m/}}~\cite{Harper15TIIS}. Ratings were discretized with a threshold of 3.5 and movies with less than 5~ratings and users with less than two movies were discarded. Inputs and outputs were built by splitting user profiles at a certain timestamp uniformly at random, ensuring a minimum of one movie in both input and output. To perform recommendations, we build on top of Wu et~al.~\cite{Wu2016} and employ a 3-layer feed-forward neural network model with 150~rectified linear units~\cite{Glorot11AISTATS} in the hidden layers. We optimize the weights of the network using cross-entropy and Adam~\cite{Kingma15ICLR}, with a learning rate of 0.001 and parameters $\beta_1=0.9$ and $\beta_2=0.999$. We evaluate the accuracy of the model using MAP.

\item Million song data set (MSD): song recommendation with the Million song data set\footnote{\url{http://labrosa.ee.columbia.edu/millionsong/tasteprofile}}~\cite{BertinMahieux11ISMIR}. We assume that a user likes a song when he/she has listened to it a minimum of 3~times. We then remove the songs that appear less than 20~times and build user profiles with a minimum of 5~songs. Inputs and outputs are split at a uniformly random timestamp. To recommend future listens to the user we use a 3-layer feed-forward neural network and 300~rectified linear units in the hidden layers. As for the rest, we proceed as with the ML task. We evaluate the accuracy of the model with MAP.

\item Amazon book reviews (AMZ): book recommendation with the Amazon book reviews data set\footnote{\url{http://jmcauley.ucsd.edu/data/amazon/}}~\cite{McAuley15KDD}. We proceed as with the ML data set, but this time setting the minimum number of ratings per book to 100. We employ a 4-layer feed-forward neural network with 300~rectified linear units in the hidden layers, and optimize its parameters with Adam. We evaluate the accuracy of the model with MAP.

\item Book crossing (BC): book recommendation with the book crossing data set\footnote{\url{http://www2.informatik.uni-freiburg.de/~cziegler/BX/}}~\cite{Ziegler05WWW}. To perform recommendations we use the same architecture and configuration as with the MSD task, but this time we use 250~units in the hidden layers.

\item YooChoose (YC): session-based recommendation with the YooChoose RecSys15 challenge data set\footnote{\url{http://recsys.yoochoose.net}}. We work with the training set of the challenge and keep only the click events. We take the first 2~million sessions of the data set which have a minimum of 2~clicks. To predict the next click, we proceed as in Hidasi et~al.~\cite{Hidasi16ICLR} and consider a GRU model~\cite{Cho14SSST}. We set the inner dimensionality to 100 and train the network with Adagrad~\cite{Duchi12JMLR}, using a learning rate of 0.01. We evaluate the accuracy of the model with RR.

\item Penn treebank (PTB): next word prediction with the Penn treebank data set~\cite{Mikolov12THESIS}. The vocabulary is limited to 10,000 words, with all other words mapped to an `unknown' token (Table~\ref{tab:data}). We consider the end of the sentence as an additional token and form input sequences of length 10. Inspired by Graves~\cite{Graves13ARXIV}, we perform next word prediction with an LSTM network~\cite{Hochreiter97NC}. We set the inner dimensionality to 250 and train the network with SGD. We use a learning rate of 0.25, a momentum of 0.99, and clip gradients to have a maximum norm of~1~\cite{Graves13ARXIV}. We evaluate the accuracy of the model with the RR of the correct prediction.

\item CADE web directory (CADE): text categorization with the CADE web directory of Brazilian web pages\footnote{\url{http://ana.cachopo.org/datasets-for-single-label-text-categorization}}~\cite{Cardoso07THESIS}. The data set contains around 40,000~documents assigned to one of 12~categories such as services, education, health, or culture. To perform classification we use a 4-layer feed-forward neural network with a softmax output. The number of units is, from input to output, 400, 200, 100, and 12, and we use rectified linear units as activations for the hidden layers. We train the network with RMSprop~\cite{Tieleman12COURSERA}, a learning rate of 0.0002, and exponential decay of 0.9. Notice that this is the only considered task where output embeddings are not required (classification into 12~text categories). We use Acc as evaluation measure.

\end{enumerate}

\subsection{Alternative approaches}
\label{sec:setup_altern}

To compare the performance of BE with the state-of-the-art, we consider 4~different embedding alternatives. We base our evaluation on performance, measured at a given input/output compression ratio. It is important to note that, in general, besides performance, alternative approaches do not present some of the other desired qualities that BE offers, such as on-the-fly operation, constant-time, no supervision, or no network/configuration changes (Secs.~\ref{sec:intro} and~\ref{sec:bloom_suitable}). Note also that methods for embedding both inputs and outputs, allowing to map embedded instances to the original ones, are scarce (Sec.~\ref{sec:relatedwork}). Because of that, in some of the considered alternatives we had to perform some adaptations.

\begin{enumerate}

\item Hashing trick (HT). We first consider the popular hashing trick for classifier and recommender inputs~\cite{Langford07TECHREP,Weinberger09ICML}. In general, these methodologies only focus on inputs and are not designed to deal with any type of output. Nonetheless, in the case of binary outputs, variants like the one used by Ganchev and Dredze~\cite{Ganchev08MLP} can be adapted to map to the original items using Eqs.~\ref{eq:like1} or~\ref{eq:like2}. In fact, considering this adaptation for recovery, the approach can be seen as a special case of BE with $k=1$ (Sec.~\ref{sec:bloom_be}).

\item Error-correcting output codes (ECOC). Originally designed for single-class targets~\cite{Dietterich95JAIR}, ECOC can be applied to class sets (inputs and outputs), with its corresponding encoding and decoding strategies~\cite{Armano12IIR}. Yet, in the case of training neural networks, it is not clear which loss function should be used. The obvious choice would be to use the Hamming distance. However, in pre-analysis, a Hamming loss turned out to be significantly inferior than cross-entropy. Therefore, we use the latter in our experiments. We construct the ECOC matrix with the randomized hill-climbing method of~\cite{Dietterich95JAIR}.

\item Pairwise mutual information (PMI). Recently, Chollet~\cite{Chollet16ARXIV} has proposed a PMI approach for embedding sets of image labels into a dense space of real-valued vectors. The approach is based on the SVD of a PMI matrix computed from counting pairwise co-occurrences. It uses cosine similarity as the loss function and, at prediction time, it performs KNN (again using cosine similarity) with the projection of individual labels to obtain a ranking. 

\item Canonical correlation analysis (CCA). CCA is a common way to learn a joint dense, real-valued embedding for both inputs and outputs at the same time~\cite{Hotelling36BM}. CCA can be computed using SVD on a correlation matrix~\cite{Hsu12JCSS} and, similarly to PMI, we can use the KNN trick to rank items or labels at prediction time. Correlation is now the metric of choice, both for the loss function and for determining the neighbors.

\end{enumerate}



\section{Results}
\label{sec:results}

\subsection{Compression and performance}

\begin{figure}[t]
    \includegraphics[width=1\linewidth,height=5.5cm]{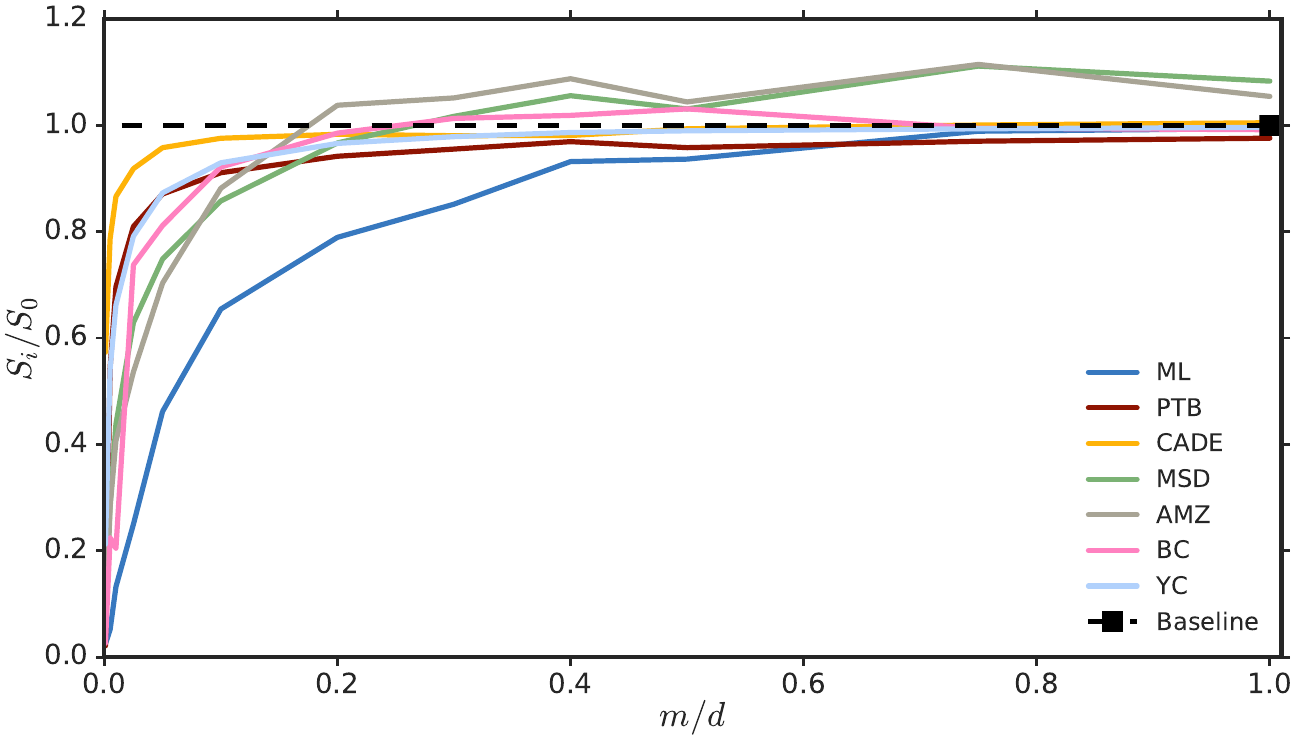}
    \caption{Score ratios $S_i/S_0$ as a function of dimensionality ratio $m/d$ using $k=4$. Qualitatively similar plots are observed for other values of $k$.}
    \label{fig:acc_m}
\end{figure}

We start by reporting on the performance of BE as a function of the embedding dimension. As mentioned, to facilitate comparisons, we report in relative terms, using score ratios $S_i/S_0$ and dimensionality ratios $m/d$. When plotting the former as a function of the latter, we see several things that are worth noting (Fig.~\ref{fig:acc_m}). Firstly, we observe that, for most of the tasks, score ratios approach 1 as $m$ approaches $d$. This indicates that the introduction of BE does not degrade the original score of the Baseline when the embedding dimension $m$ is comparable to the original dimension $d$. Secondly, we observe that the lower the dimensionality ratio, the lower the score ratio. This is to be expected, as one cannot embed sets of items with their intrinsic dimensionality to an infinitesimally small $m$. Importantly, the reduction of $S_i/S_0$ should not be linear with $m/d$, but should maximize $S_i$ for low $m$ (thus getting curves close to the top left corner of Fig.~\ref{fig:acc_m}). We see that BE fulfills this requirement. In general, we can reduce the input/output size up to 5~times ($m/d=0.2$) and still maintain more than 92\% of the value of the original score. The ML task is the only exception, which we think is due to the abnormally high density of the data (Table~\ref{tab:data}), inhibiting the embedding to low dimensions\footnote{Note that the ML data is essentially collected through a survey-type method~\cite{Harper15TIIS}.}. CADE is the task for which BE achieves the highest $S_i$ for low $m$. Presumably, the CADE task is the easiest one we consider, as only input embeddings are required. 

An additional observation is worth noting (Fig.~\ref{fig:acc_m}). Interestingly, we find that BE can improve the scores over the Baseline for a number of tasks. That is the case for 3 out of the 7 considered tasks (all of them recommendation tasks): MSD with $m/d\geq 0.3$, AMZ with $m/d\geq 0.2$, and BC with $0.3\leq m/d\leq 0.6$. The fact that an embedding performs better than the original Baseline has been also observed in some other methods for specific data sets~\cite{Weston02NIPS,Langford07TECHREP,Chollet16ARXIV}. For instance, Chollet~\cite{Chollet16ARXIV} has reported increases up to 7\% using the PMI approach on the so-called JFT data set. Here, depending on the task and the embedding dimension, relative increases go from 1~to~12\%. Given that the data sets where we observe these increases are some of the less dense ones (Table~\ref{tab:data}), we hypothesize that, in the case of BE, such increases come from having $k$ times more active positions in the ground truth output (recall that one output item is projected $k$ times using $k$ independent hash functions, Sec.~\ref{sec:bloom_be}). With $k$ more times elements set to 1 in the output, a better estimation of the gradient may be computed (larger errors that propagate back to the rest of the network).

\begin{figure}[t]
    \includegraphics[width=0.49\linewidth,height=4cm]{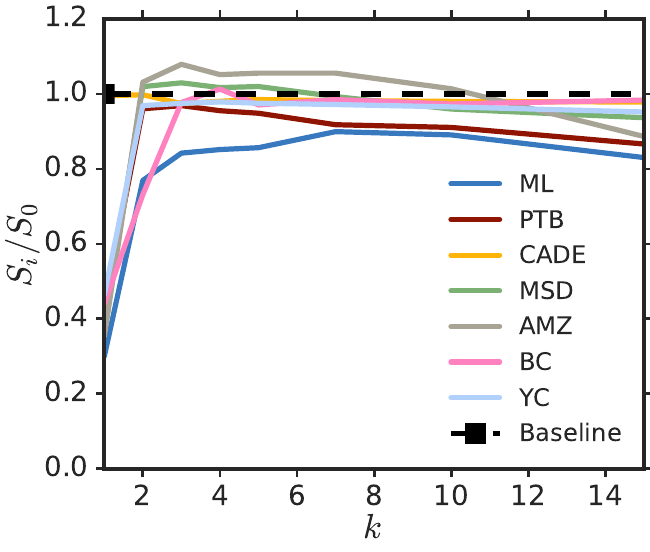} \includegraphics[width=0.49\linewidth,height=4cm]{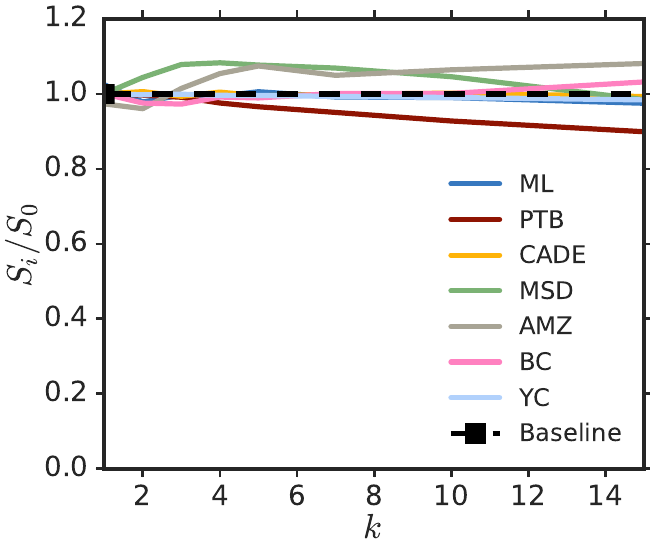}
    \caption{Score ratios $S_i/S_0$ as a function of the number of hash functions $k$: using $m/d=0.3$ (left) and $m/d=1$ (right).}
    \label{fig:acc_k}
\end{figure}

We now focus on performance as a function of the number of projections $k$ (Fig.~\ref{fig:acc_k}), reporting score ratios $S_i/S_0$ as above. From repeating the plots for different values of $m/d$, we observe that $S_i/S_0$ is always low for $k=1$ (Fig.~\ref{fig:acc_k}, left), except when $m$ approaches $d$, where we have an almost flat behavior (Fig.~\ref{fig:acc_k}, right). In general, $S_i/S_0$ jumps up for $k\geq 2$ and remains stable until $k\approx 10$, where the decrease of $S_i/S_0$ becomes more apparent (Fig.~\ref{fig:acc_k}, left). The best operating range typically corresponds to $2\leq k\leq 4$. The ML task is again an exception, with a best operating range around $7\leq k\leq 10$.

\subsection{Training and retrieval time}

\begin{figure}[t]
    \includegraphics[width=0.49\linewidth,height=4.5cm]{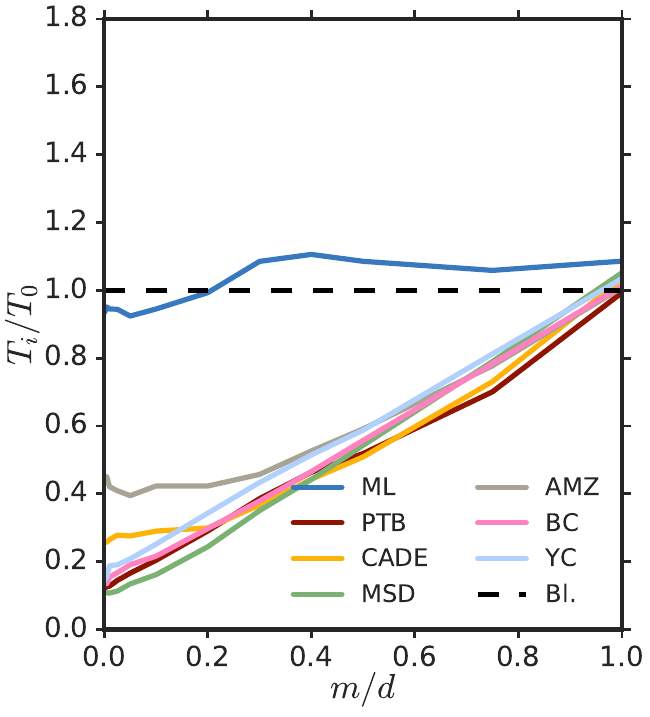} \includegraphics[width=0.49\linewidth,height=4.5cm]{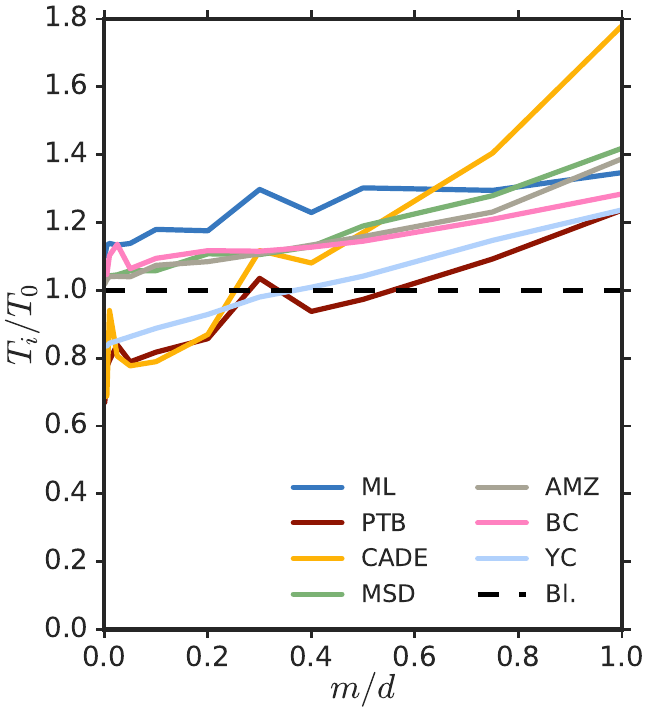}
    \caption{Time ratios $T_i/T_0$ as a function of dimensionality ratios $m/d$ with $k=4$: training time (left) and evaluation time (right). Qualitatively similar plots are observed for other values of $k$. Bl.~denotes baseline.}
    \label{fig:time}
\end{figure}

Besides performance scores, it is interesting to assess whether the reduction of input and output dimensions has an effect to training and evaluation times. To this end, we plot the time ratios $T_i/T_0$ as a function of the dimensionality ratio $m/d$ (Fig.~\ref{fig:time}). Regarding training times, we basically observe a linear decrease with $m/d$ (Fig.~\ref{fig:time}, left). ML is an exception to the trend, and CADE and AMZ experiment almost no decrease for very low dimensionality ratios $m/d<0.2$. In general, we confirm faster training times thanks to the reduction of the number of parameters of the model, dominated by input/output matrices (output dimension also affecting the time to compute the loss function). We obtain a 2~times speedup for a 2~times input/output compression and, roughly, a little bit over 3~times speedup for a 5~times input/output compression. Regarding evaluation times, we also observe a linear trend (Fig.~\ref{fig:time}, right). However, this time, $T_i/T_0$ is not as low, with values slightly above 1 but always below 1.5 (with the exception of CADE for $m/d>0.6$). Overall, this indicates that, compared to the Baseline evaluation time, the mapping used by BE when reconstructing the output does not introduce an overwhelming amount of extra computation time. With the exception of ML, extra computation time is below 20\% for $m/d< 0.5$.

\subsection{Comparison with alternatives}

Finally, we compare the performance of BE to the one of the considered alternative methods. We do so by establishing a dimensionality ratio $m/d$ and computing the corresponding score ratio $S_i/S_0$ for a given task (Table~\ref{tab:compare}). We see that BE is better than the alternative methods in 5~out of the 7 tasks (10 out of the 14 considered test points). PMI is better in one of the tasks (CADE) and CCA is better also in one of the tasks (AMZ). It is relevant to note that, when BE wins, it always does so by a relatively large margin (see, for instance, the ML or YC tasks). Otherwise, when an alternative approach wins, generally it does so by a smaller margin (see, for instance, the AMZ task). 

These results become more relevant if we realize that PMI and CCA are both SVD-based approaches, introducing a separate degree of supervised learning to the task, exploiting pairwise item co-occurrences and correlations, respectively (Sec.~\ref{sec:setup_altern}). In contrast, BE does not require any learning. We formulate a co-occurrence-based version of BE below, which achieves moderate performance increments over BE and more closely approaches the performance of PMI and CCA on the two tasks where BE was not already performing best. A further interesting thing to note is that we confirm the small variation in the score ratios obtained for $2\leq k\leq 10$ (Fig.~\ref{fig:acc_k}). Here, score ratios for $3\leq k\leq 5$ are often comparable in a statistical significance sense (Table~\ref{tab:compare}).

\begin{table*}[t]
    \caption{Comparison of BE with the considered alternatives. Score ratios $S_i/S_0$ for different combinations of data set and compression ratio $m/d$. Best results are highlighted in bold, up to statistical significance (Mann-Whitney~U, $p\!>\!0.05$).}
    \label{tab:compare}
    \setlength{\tabcolsep}{10pt}
    \begin{tabular}{lc|cccc|ccc}
        \toprule
        \multicolumn{2}{c|}{Test point} & \multicolumn{4}{c|}{Alternative methods}  & \multicolumn{3}{c}{BE} \\
        Data set    & $m/d$ & HT    & ECOC  & PMI   & CCA   & $k=3$ & $k=4$ & $k=5$ \\
        \midrule
        ML          & 0.2     & 0.234 & 0.342 & 0.043 & 0.209 & \bf 0.750 & \bf 0.770 & 0.722 \\
        ML          & 0.3     & 0.285 & 0.208 & 0.045 & 0.200 & 0.796 & \bf 0.813 & \bf 0.815 \\
        PTB         & 0.2     & 0.357 & 0.453 & 0.837 & 0.638 & \bf 0.919 & 0.908 & 0.881 \\
        PTB         & 0.4     & 0.528 & 0.454 & 0.836 & 0.695 & \bf 0.942 & 0.920 & 0.902 \\
        CADE        & 0.01    & 0.857 & 0.359 & \bf 0.984 & 0.928 & 0.862 & 0.853 & 0.855 \\
        CADE        & 0.03    & 0.914 & 0.363 & \bf 1.002 & 0.950 & 0.914 & 0.925 & 0.926 \\
        MSD         & 0.05    & 0.078 & 0.268 & 0.216 & 0.679 & 0.695 & \bf 0.738 & \bf 0.738  \\
        MSD         & 0.1     & 0.151 & 0.310 & 0.321 & 0.740 & \bf 0.835 & \bf 0.841 & 0.832  \\
        AMZ         & 0.1     & 0.166 & 0.182 & 0.851 & \bf 1.030 & 0.864 & 0.881 & 0.861 \\
        AMZ         & 0.2     & 0.289 & 0.185 & 0.995 & \bf 1.048 & 1.016 & 1.029 & 1.008  \\
        BC          & 0.05    & 0.189 & 0.817 & 0.022 & 0.313 & 0.777 & 0.750 & \bf 0.837  \\
        BC          & 0.1     & 0.199 & 0.886 & 0.025 & 0.465 & \bf 0.965 & \bf 0.919 & 0.831  \\
        YC          & 0.03    & 0.150 & 0.076 & 0.776 & 0.466 & \bf 0.841 & \bf 0.858 & \bf 0.858  \\
        YC          & 0.05    & 0.240 & 0.083 & 0.777 & 0.517 & \bf 0.919 & \bf 0.910 & \bf 0.928  \\
        \bottomrule
    \end{tabular}
\end{table*}


\section{Going one step further with co-occurrence-based collisions}
\label{sec:coobloom}

\subsection{Co-occurrence-based embedding}

In Bloom filters and BE, collisions are unavoidable due to the lower embedding dimensionality and the use of multiple projections (Sec.~\ref{sec:bloom}). In addition, we have seen that some alternative approaches produce embeddings by exploiting co-occurrence information (Secs.~\ref{sec:relatedwork} and~\ref{sec:setup_altern}). Here, we study a variant of BE that takes advantage of co-occurrence information to adjust the collisions that will inevitably take place when performing the embedding. We denote this approach by co-occurrence-based Bloom embedding (CBE).

What we propose is a quite straightforward approach to CBE, which does not add much extra pre-computation time. Training and testing times remain the same, as CBE uses a pre-computed hashing matrix $\ma{H}$ (Sec.~\ref{sec:bloom_be}). The general idea of the proposed approach is to `re-direct' the collisions of the co-occurring items to the same bits or positions of $\ve{u}$. Our implementation of this idea is detailed in Algorithm~\ref{alg:cbe}, and briefly explained below.

\begin{algorithm}[t]
	\caption{Pseudocode for CBE.}
	\label{alg:cbe}
	\begin{algorithmic}[1]
		\vspace*{0.1cm}
		\REQUIRE Input and/or output instances $\ma{X}$ ($n\times d$ sparse binary matrix), embedding dimensionality $m$, number of projections $k$, and pre-computed hashing matrix $\ma{H}$ ($d\times k$ integers matrix). \\
		\ENSURE Co-occurrence-based hashing matrix $\ma{H}'$. \\
		\vspace*{0.2cm}
        \STATE $\ma{C}\leftarrow \ma{X}^{\text{T}}\ma{X}$ \\
        \STATE $\ma{C}\leftarrow \ma{C}$ $\odot$ \textsc{sgn}$( \ma{C}-$\textsc{avgfreq}$(\ma{X}))$ \\
        \STATE $\ve{c}^{\textsc{val}},\ve{c}^{\textsc{row}},\ve{c}^{\textsc{col}} \leftarrow$ \textsc{coord}$($\textsc{lowtri}$(\ma{C}))$ \\
        \FOR{$i$ \textbf{in} \textsc{argsort}$(\ve{c}^{\textsc{val}})$}
            \STATE $a,b\leftarrow c^{\textsc{row}}_i,c^{\textsc{col}}_i$ \\
            \STATE $r \leftarrow$ \textsc{urnd}$(1,m,\se{h}_a\cup\se{h}_b)$ \\
            \STATE $j_a \leftarrow$ \textsc{urnd}$(1,k,\emptyset)$ \\
            \STATE $j_b \leftarrow$ \textsc{urnd}$(1,k,\emptyset)$ \\
            \STATE $h_{a,j_a}, h_{b,j_b} \leftarrow r$ \\
        \ENDFOR
        \RETURN $\ma{H}$\\
	\end{algorithmic}
\end{algorithm}

First, we count pairwise co-occurrences and store them in a sparse matrix $\ma{C}$ (line~1). Next, we threshold $\ma{C}$ by the average item frequency in $\ma{X}$ using the Hadamard product $\odot$ and a component-wise sign function (line~2). We then get the lower triangular part of $\ma{C}$ and return it in coordinates format, that is, using a tuple of values, row indices, and column indices (line~3). We will use the order in $\ve{c}^{\textsc{val}}$ to update the hash matrix $\ma{H}$. To do so, we first loop over the indices of the sorted values of $\ve{c}^{\textsc{val}}$ in increasing order (line~4). After selecting the corresponding items $a$ and $b$ (line~5), we then draw integers from \textsc{urnd} (lines~6--8). The function \textsc{urnd}$(x,y,\se{z})$ is a uniform random integer generator between $x$ and $y$ (both included) such that the output integer is not included in the set $\se{z}$, that is, \textsc{urnd}$(x,y,\se{z})\not\in\se{z}$. Rows $a$ and $b$ of $\ma{H}$ are transformed to sets $\se{h}_a$ and $\se{h}_b$ and its union is computed (line~6). Finally, we use the integers generated by \textsc{urnd} to pick projections $j_a$ and $j_b$ from $\ma{H}$, and assign them the same bit $r$ (line~9). By updating the projections in $\ma{H}$ in increasing order of co-occurrence (line~4), we give priority to the pairs with largest co-occurrence, setting them to collide to the same bit $r$ (line~9).

\subsection{CBE results}

\begin{figure}[t]
    \includegraphics[width=1\linewidth]{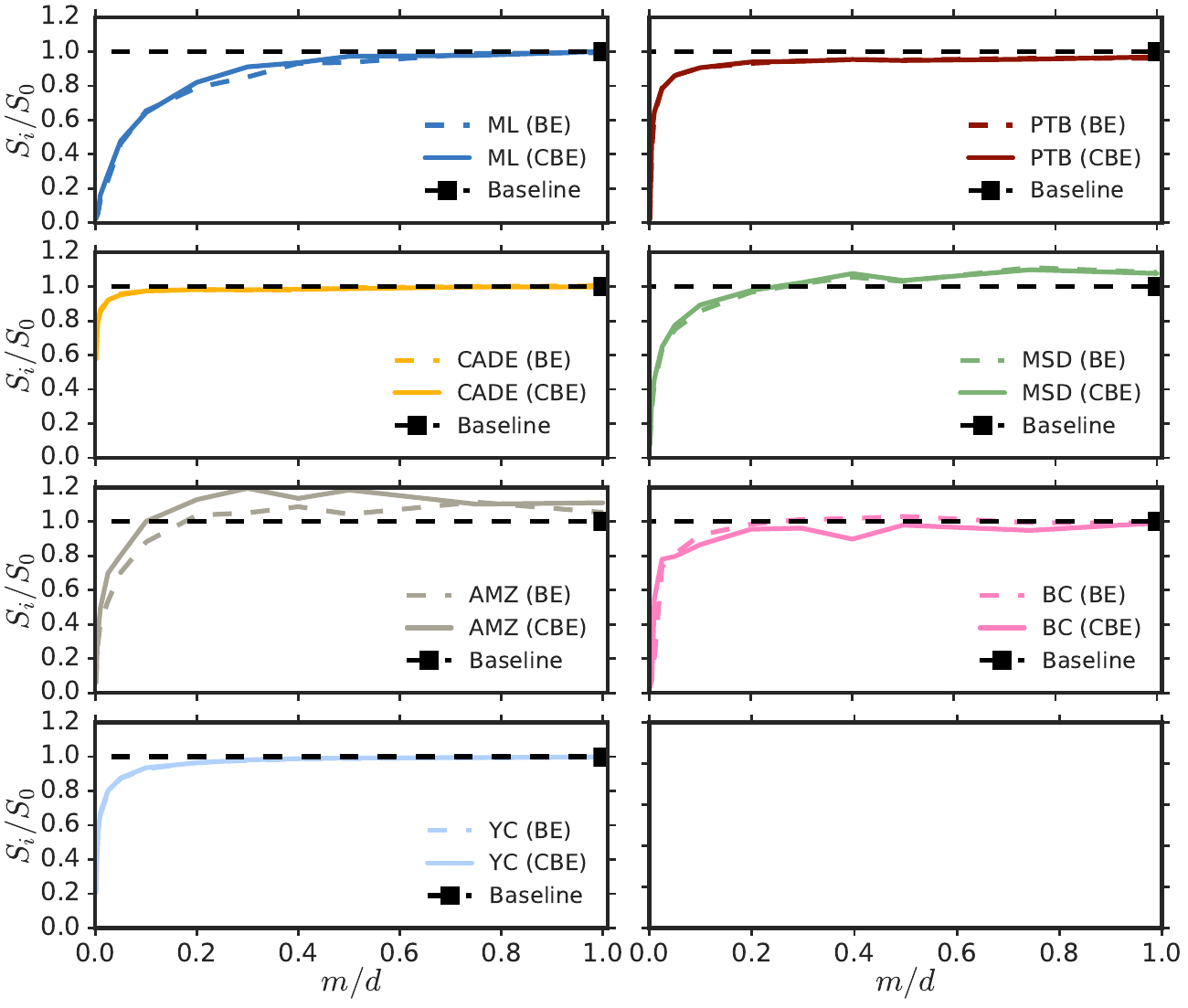}
    \caption{Comparison of score ratios $S_i/S_0$ as a function of dimensionality ratio $m/d$ for BE (dashed lines) and CBE (solid lines) using $k=4$. Qualitatively similar plots are observed for other values of $k$.}
    \label{fig:acc_comp}
\end{figure}

Overall, the performance of CBE only provides moderate increments over the original BE approach (Fig.~\ref{fig:acc_comp}). With the exception of the BC task, the performance of CBE is always higher than the one of BE. However, with the exception of the AMZ task, we do not observe dramatic increases of CBE over BE. On average, such increases are between 0.4\% and 8.4\% (Table~\ref{tab:acc_comp}, right). One possible explanation for these moderate performance increases is the low co-occurrence in the considered data (Table~\ref{tab:acc_comp}, left). As it can be seen, typically less than 3\% of all possible pairs show a co-occurrence. Moreover, the average co-occurrence count of such co-occurring pairs is very low, with ratios $\rho$ to the total number of instances $n$ in the order of $10^{-5}$ or $10^{-6}$.

\begin{table}[t]
    \caption{Co-occurrence statistics and average score increase of CBE over BE. From left to right: data set name, input percent of co-occurrent pairs (\%), input average co-occurrence ratio of co-occurrent pairs ($\rho$), output percent of co-occurrent pairs (\%), output average co-occurrence ratio of co-occurrent pairs ($\rho$), and average score increases of CBE over BE (\%; calculated using $100(S_j-S_i)/S_0$ and averaging over all $m/d$ points). Co-occurrence values for PTB and YC inputs correspond to considering training sequences, not isolated sequence items.}
    \label{tab:acc_comp}
    \begin{tabular}{lcccc|cc}
        \toprule
        Data set~~ & \multicolumn{4}{c|}{Co-occurrence statistics} & \multicolumn{2}{c}{Score increase} \\
                & \multicolumn{2}{c}{Input} & \multicolumn{2}{c|}{Output} & \multicolumn{2}{c}{(\%)} \\
                & \% & $\rho$ & \% & $\rho$ & $k=3$ & $k=4$ \\
        \midrule
        ML      & 25.2 & $1.3\cdot 10^{-4}$ & 32.9 & $1.0\cdot 10^{-4}$ & $+0.9$ & $+1.7$ \\
        PTB     & 3.3 & $2.4\cdot 10^{-5}$ & 0 & 0 & $+0.1$ & $+0.9$ \\
        CADE    & 1.3 & $8.8\cdot 10^{-5}$ & N/A & N/A & $-0.4$ & $-0.1$ \\
        MSD     & 1.3 & $3.0\cdot 10^{-6}$ & 1.3 & $3.1\cdot 10^{-6}$ & $+0.5$ & $+1.5$ \\
        AMZ     & 3.0 & $1.8\cdot 10^{-6}$ & 3.0 & $1.8\cdot 10^{-6}$ & $+6.6$ & $+8.4$ \\
        BC      & 0.8 & $4.9\cdot 10^{-5}$ & 0.4 & $4.9\cdot 10^{-5}$ & $-3.4$ & $-1.0$ \\
        YC      & 0.2 & $1.5\cdot 10^{-6}$ & 0 & 0 & $+0.4$ & $+0.3$ \\
        \bottomrule
    \end{tabular}
\end{table}

Despite being moderate on average, we observed that the increments provided by CBE were more prominent for low dimensionality ratios $m/d$. By relating CBE with the best approaches resulting from the previous comparison of Table~\ref{tab:compare}, we see that CBE is generally better than BE, sometimes with a statistically significant difference (Table~\ref{tab:compare_cbe}). Furthermore, we see that CBE, being based on co-occurrences, more closely approaches PMI and CCA in the tasks where those were performing best, and even outperforms them in one test point (AMZ, $m/d=0.2$). Being closer to those co-occurrence-based approaches is an indication that CBE leverages co-occurrence information to some extent.

\begin{table}[t]
    \caption{Comparison of CBE versus the results in Table~\ref{tab:compare}. Score ratios $S_i/S_0$ for different combinations of data set and compression ratio $m/d$. Best results are highlighted in bold, up to statistical significance (Mann-Whitney-U, $p\!>\!0.05$).}
    \label{tab:compare_cbe}
    \setlength{\tabcolsep}{9pt}
    \begin{tabular}{lc|cc|cc}
        \toprule
        \multicolumn{2}{c|}{Test point} & \multicolumn{2}{c|}{Best so far}  & \multicolumn{2}{c}{CBE}\\
        Data set    & $m/d$   & Method & $S_i/S_0$ & $k=3$ & $k=4$ \\
        \midrule
        ML          & 0.2     & BE & 0.770 & 0.760 & \bf 0.781 \\
        ML          & 0.3     & BE & 0.815 & 0.812 & \bf 0.867 \\
        PTB         & 0.2     & BE & \bf 0.919 & \bf 0.915 & \bf 0.907 \\
        PTB         & 0.4     & BE & \bf 0.942 & \bf 0.937 & 0.922 \\
        CADE        & 0.01    & PMI & \bf 0.984 & 0.854 & 0.853 \\
        CADE        & 0.03    & PMI & \bf 1.002 & 0.921 & 0.922 \\
        MSD         & 0.05    & BE & 0.738 & \bf 0.759 & 0.756 \\
        MSD         & 0.1     & BE & 0.841 & 0.856 & \bf 0.873 \\
        AMZ         & 0.1     & CCA & \bf 1.030 & 0.994 & 0.991 \\
        AMZ         & 0.2     & CCA & 1.048 & \bf 1.109 & \bf 1.117 \\
        BC          & 0.05    & BE & \bf 0.837 & 0.774 & 0.808 \\
        BC          & 0.1     & BE & \bf 0.965 & 0.880 & 0.878 \\
        YC          & 0.03    & BE & 0.858 & 0.871 & \bf 0.880 \\
        YC          & 0.05    & BE & 0.928 & 0.933 & \bf 0.936 \\
        \bottomrule
    \end{tabular}
\end{table}


\section{Conclusion}
\label{sec:conclusion}

We have proposed the use of Bloom embeddings to represent sparse high-dimensional binary-coded inputs and outputs. We have shown that a compact representation can be obtained without compromising the performance of the original neural network model or, in some cases, even increasing it by a substantial factor. Due to the compact representation, the loss function and the input and output layers deal with less parameters, which results in faster training times. The approach compares favorably with respect to the considered alternatives, and offers a number of further advantages such as on-the-fly operation or zero space requirements, all this without introducing changes to the core network architecture, task configuration, or loss function.

In the future, besides continuing to exploit co-occurrences, one could enhance the proposed approach by considering further extensions of Bloom filters such as counting Bloom filters~\cite{Bonomi06ESA}. In theory, those extensions could provide a more compact representation by breaking the binary nature of the embedding. However, they could require the modification of the loss function or the mapping process (Eqs.~\ref{eq:like1} and~\ref{eq:like2}). A faster mapping process using the sorted probabilities of $\ve{v}$ could also be studied. A detailed, comparative analysis of false positives and false negatives is also pending. Finally, it would be interesting to assess the utility of BE in combination with classical collaborative filtering approaches or factorization machines~\cite{Rendl10ICDM}.


\begin{acks}
We thank the curators of the data sets used in this study for making them publicly-available\iftoggle{reveal_id}{ and Santi Pascual for his comments on a previous version of the paper.}{.}
\end{acks}


\clearpage

\bibliographystyle{ACM-Reference-Format}
\bibliography{bibjoan,bibalex} 

\end{document}